\documentclass[hyphenate]{cascadilla}

\usepackage[T1]{fontenc}
\usepackage[utf8]{inputenc}

\usepackage{synttree}
\branchheight{4ex}

\usepackage[x11names,rgb]{xcolor}
\usepackage{tikz}
\usetikzlibrary{snakes,arrows,shapes}

\usepackage{amsmath}
\usepackage{amssymb}

\usepackage{tipa}
\usepackage{lingmacros}
\usepackage{url}
\makeatletter
\def\url@leostyle{%
  \@ifundefined{selectfont}{\def\UrlFont{\sf}}{\def\UrlFont{\small\ttfamily}}}
\makeatother
\title{Acquisition of morphological families and derivational series from a machine readable dictionary}
\author{Nabil Hathout}

\institution{Université de Toulouse}
\completed{\today}
\copyrightyear{2009}

\begin{document}

\maketitle

The question we address in this paper is: how to perform morphological analysis in the framework of word-based morphology, that is without resorting to the notions of morpheme, affix, morphological exponent or any representation of these concepts?  We do not present here  a fully fledged answer, but we describe a general framework for doing so and a method for computing a large part of the intended analysis.  The paper is divided into five parts.  In section \ref{sec:toward-word-based}, we outline the objectives of the research and the method.  We then detail the measure of morphological similarity (section~\ref{sec:morph-relatedness}), and, the formal analogy we use to filter the morphological neighborhoods (section~\ref{sec:analogy}).  We then present some preliminary results (section~\ref{sec:first-results}) and a short conclusion (section~\ref{sec:future-prospects}).

\section{Toward a computational word-based morphology}
\label{sec:toward-word-based}

\subsection{Word-based vs morpheme-based morphology}
\label{sec:word-morpheme-based-morphology}

\begin{figure}[h]
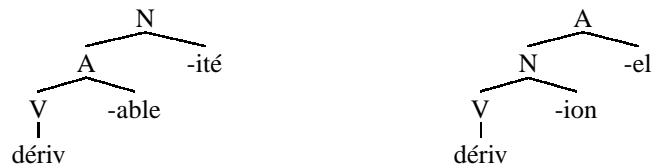

\mbox{}\hfill
\synttree[N [A [V [dériv]] [-able]] [-ité]]
\hfill
\synttree[A [N [V [dériv]] [-ion]] [-el]]
\hfill\mbox{}
    \caption{Word structure of the French noun \emph{dérivabilité} `derivability' and the French adjective \emph{dérivationnel} `derivational' in morpheme-based morphology.}
  \label{fig:word-structure}
\end{figure}

In standard morpheme-based morphology, words are made up of morphemes.  The morphemes are combined by rules of inflection, derivation and composition.  They have structures which are usually represented as trees like the ones in figure~\ref{fig:word-structure}.  Morpheme-based morphology is both elegant and easy to use, but it suffers from many drawbacks \citep{anderson92.a-morphous-morphology,aronoff94.morphology-itself}; there is no need to enumerate them here.

In word-based morphology \citep{aronoff76.word-formation,bybee88.lexical-organization,bybee95.regular-morphology,neuvel2001.vive-difference,burzio99.surface-to-surface-morphology,blevins2006.word-based}, the minimal units are the words.  Therefore, they do not have any structure.  Morphological structure then becomes a level of organization of the lexicon, made up of the morphological relations that hold between the words.  Some of them play a special role, namely the relations between the words that belong to the same lexeme, to the same inflectional series, to the same morphological family and to the same derivational series.  These four types of aggregates can be illustrated by the lexeme and the inflectional series of the French verb form \emph{dérivons} `derive' and by the morphological family and the derivational series of the deverbal noun \emph{dérivation}:
\begin{itemize}
\item the \textbf{lexeme} of \emph{dérivons} contains the inflected forms of the verb \emph{dériver} `derive': \emph{dériver}, \emph{dérive}, \emph{dériverez}, \emph{dérivaient}, \emph{dérivées}, \emph{dérivions}, etc.;
\item \emph{dérivons} belongs to an \textbf{inflectional series} of first person plural, present indicative verb forms which also contains \emph{acclimatons} `acclimate', \emph{compilons} `compile', \emph{éduquons} `educate', \emph{localisons} `localize', \emph{varions} `vary', etc.;
\item the \textbf{morphological family} of \emph{dérivation} contains words such as \emph{dériver}, \emph{dérivable}, \emph{dérivatif} `derivative', \emph{dérivationnel}, \emph{dérivabilité}, \emph{dérive} `drift', \emph{dériveur} `sailing dinghy', etc.;
\item \emph{dérivation} belongs to a \textbf{derivational series} of deverbal nouns in \emph{-ion} such as \emph{acclimatation} `acclimation', \emph{compilation}, \emph{éducation}, \emph{localisation}, \emph{variation}, etc.
\end{itemize}
In the rest of the paper, we concentrate only on the derivational part of the morphological structure.

Notice that morphology does not reduce to this lexical structure.  For instance, \emph{anti universal healthcare} in (\ref{ex:anti-universal-healthcare}) is a morphological construct\footnote{%
\emph{Construct} is used in this paper as a generic term to designate any linguistic object produced by the morphology.%
} that is not likely to enter the lexicon nor have a place in the structure.
\enumsentence{
  \label{ex:anti-universal-healthcare}
  \emph{All those anti feminist, anti Democrat, anti giving everyone the right to vote, \textbf{anti universal healthcare}, anti all kinds of things I thought no one was anti.}\\
  \url{www.talkleft.com/story/2008/4/18/204142/362}
}
Actually, lexicon and morphology must be clearly separated: the main function of the lexicon is to memorize and organize the words that a speaker knows; the main function of the morphology is to produce and analyze words.  The constructs produced by the morphology are designed to enter the lexicon and extend the lexical structure. In return, the lexicon provides the morphology with  the information it needs to create and analyze morphological constructs.  However, the distribution of morphological information between lexicon and morphology varies according to morphological theories.  In morpheme-based morphology, each word has a separate structure, the lexicon is just a bag of morphemes (and possibly of fully analyzed words) and morphological rules encode the bulk of the morphological information.  In word-based morphology, the distribution of the information is more even.  The lexicon contains both the words and the morphological relations that hold between them.  Morphology is made up of processes that extend the existing lexical structure with new words.  These processes can also be used to create constructs such as (\ref{ex:anti-universal-healthcare}) that have no place in the lexicon.  In this paper, we are concerned only with the lexicon structure.

The morpheme-based \emph{vs} word-based distinction shows up on the computational level. In the morpheme-based conception, the morphological analysis of a word aims at segmenting it into a sequence of morphemes \citep{dejean98.PaGNLL,gaussier99.ACL,schone2000.CoNLL,goldsmith2001.CL,creutz2002.ACL,bernhard2006.FinTAL}.  For instance, \emph{dérivation} is analyzed as made up of two segments \emph{dérivat} and \emph{-ion}, the first being identified as the root morpheme \emph{dériv} and the second as the suffix \emph{-ion}.  In a word-based approach, the aim of a morphological analysis is to discover the relations between the word and the other lexical items and to identify its morphological family and its derivational series.  For instance, an analysis of the French word \emph{dérivation} is satisfactory if it connects \emph{dérivation} with enough members of its morphological family (\emph{dériver}, \emph{dérivationnel}, \emph{dérivable}, etc.) and its derivational series (\emph{formation} `education', \emph{séduction}, \emph{émission}, \emph{vision}, etc.).

The morphological relations are organized into analogical series.  For instance, the relation between \emph{dérivation} and \emph{dérivable} is the basis of analogies such as \emph{dérivation}:\emph{dérivable} :: \emph{variation}:\emph{variable},\footnote{%
  The notation $a:b::c:d$ is used as a shorthand for the statement that $(a,b,c,d)$ forms an analogical quadruplet, or in other words that $a$ is to $b$ as $c$ is to $d$.%
} \emph{dérivation}:\emph{dérivable} :: \emph{modification}:\emph{modifiable}, \emph{dérivation}:\emph{dérivable} :: \emph{adaptation}:\emph{adaptable}, \emph{dérivation}:\emph{dérivable} :: \emph{observation}:\emph{observable}, etc.  Similarly, the relation between \emph{dérivation} and \emph{variation} gives rise to a series of analogies such as \emph{dérivation}:\emph{variation} :: \emph{dériver}:\emph{varier}, \emph{dérivation}:\emph{variation} :: \emph{dérivationel}:\emph{variationnel}, \emph{dérivation}:\emph{variation} :: \emph{dérivabilité}:\emph{variabilité}, \emph{dérivation}:\emph{variation} :: \emph{dérivable}:\emph{variable}, etc.  These examples show how morphological analogies connect the morphological families and the derivational series.

\subsection{Combining morphological relatedness and formal analogy}
\label{sec:comb-moph-analogy}

In the remainder of the paper, we present a computational model that makes the morphological derivational structure of the lexicon emerge from the semantic and the formal regularities of the words it contains.  A first experiment is currently underway on the lexicon of French using the \emph{Trésor de la Langue Française informatisé}  machine readable dictionary (or \emph{TLFi} for short; \texttt{atilf.atilf.fr/tlf.htm}).  Our aim  is to create a lexicon that provides the morphological family and the derivational series of the words it contains.  This morphological lexicon owes its strength to the global description of a significant part of the French lexicon.  We are building it  from a lexicographical resource because we need semantic descriptions for a large number of words.  We are fully aware of the limitations of the lexicographical descriptions but the benefits of using dictionaries far exceed them.  Dictionaries provide definitions and graphemic / phonological representations for a significant part of the lexicon.  Besides, lexicographic descriptions are more easy to use than data extracted from corpora since they only present the sub-senses and the definitions of the most representative usages of the words.

Our method relies on a measure of morphological relatedness that brings the members of morphological families and derivational series closer.  This measure takes into account both the formal and the semantic similarities between the words.  The method also relies on the discovery of formal analogies among morphological neighbors.  The use of analogy is quite common in computational morphology \citep{skousen89.analogical-modeling,lepage98.COLING,van-den-bosch99.morphological-analysis,pirrelli99.hidden-dimension,hathout2005.CahLex,stroppa2005.CoNLL}.  The main novelty of the method is to combine it with a measure of morphological relatedness.  First, lexical similarity is used in order to select quadruplets of words that are related to each other.  The candidates are then checked by means of analogy.  The two techniques are complementary.  Morphological similarity can be computed for large numbers of words, but it is too coarse-grained to discriminate between the words that are actually morphologically related and the ones that are not.  Formal analogy is then used to perform fine grained filtering but is costly to calculate.

More generally, our approach is original in that:
\begin{enumerate}

\item The computational model is purely word-based.  The discovery of morphological relations between words do not involve the notions of morpheme, affix, morphological exponent, etc.\ or any representation of these concepts.

\item Membership in families and series is gradient.  It accounts, for instance, for the fact that \emph{dériveur} is morphologically and semantically closer to \emph{dérive} than to \emph{dérivationnellement} `derivationally', even if the three words belong to the same family.  The model connects the words that share semantic and / or formal features.  The more features they share and the more specific these features are, the closer the words are.

\item It implements the theoretical proposals of \cite{bybee88.lexical-organization,bybee95.regular-morphology} and \cite{burzio99.surface-to-surface-morphology} in a straightforward manner.

\item It is efficient enough to be used to build a large morphological resource semi-automatically.

\end{enumerate}
Besides,  the model integrates semantic and formal information in a uniform manner.  All kinds of semantic information (lexicographic definitions, synonyms, synsets, etc.) and formal information (phonological, graphemic, syllabic, etc.) can be used.  These specifications can be cumulated easily in spite of differences in nature and origin.  The model takes advantage of the redundancy of the features and is fairly insensitive to variation and exceptions.  It is robust and language independent.

Technically, the model joins:
\begin{enumerate}

\item the representation of the lexicon as a graph and its exploration through random walks, along the lines of \citep{gaume2002.JETAI,gaume2005.hierarchy-lexical-organization,muller2006.textgraphs}, and

\item formal analogies on words \citep{lepage98.COLING,lepage2003.HDR,stroppa2005.CoNLL,langlais2007.EMNLP-CoNNL}. This approach does not make use of morphemes.  Correspondence between words is calculated directly on their graphemic representations.

\end{enumerate}

\subsection{Network lexicon}
\label{sec:network-lexicon}

The morphological lexicon we intend to build is a network of words with connections mainly defined by the morphological families and the derivational series.  This primary structure is completed with a set of analogies between pairs of morphologically related lexemes and with a morphological distance.  The resulting lexicon is remarkably flexible and can adequately represent various morphological phenomena.  One of them is allomorphy, which corresponds to locations in the network where there is a mismatch between the formal analogies and the organization into families and series. For instance, the French deverbal noun \emph{dénivellation} `unevenness' can be identified as an allomorphic form because (\emph{i}) it belongs to a series of words ending in \emph{-ion}, (\emph{ii}) it is a member of the family of the verb \emph{déniveler} `make uneven' and, (\emph{iii}) it is morphologically the closest noun to this verb.  Nouns in \emph{-ion} and more specifically in \emph{-ation} are normally involved in  analogies with their closest verbs such as \emph{dérivation}:\emph{dériver} :: \emph{compilation}:\emph{compiler}.  The absence of such analogies for \emph{dénivellation} appears as a gap in the analogical grid.  This gap is the sign of an allormorphy.  Another cue is the near identity of \emph{dénivellation} with the string \emph{dénivelation} which would have allowed \emph{déniveler} to enter the main set of analogies involving the nouns ending in \emph{-ion} (i.e. the set of analogies with the strongest morphological density).

The lexicon also accounts for the similarity and difference between \emph{curieux} `curious' and \emph{furieux} `furious' in the same way \citep{jackendoff75.morphological-regularities}.  On the one hand, \emph{furieux} can be analyzed as an adjective derived from the noun \emph{furie} `fury' but we cannot do so for \emph{curieux} since it is no longer semantically related to \emph{cure} `care'.  On the other hand, both adjectives have the formal and the semantic features of \emph{-eux} derivatives. In the lexicon we propose, both adjectives belong to the same derivational series.  On the other hand, \emph{furieux} and \emph{furie} participate in a series of analogies with \emph{mélodieux}:\emph{mélodie} `melodious':`melody', \emph{harmonieux}:\emph{harmonie} `harmonious':`harmony', \emph{facétieux}:\emph{facétie} `facetious':`joke', etc.\ while \emph{curieux} does not.  This example shows the flexibility of our model and the higher descriptive precision we obtain from the derivational series and the morphological analogies.  By contrast, the similarity of \emph{curieux} with \emph{furieux} cannot be described in a morphematic model or in any model lacking derivational series.  Note that the term ``derivational series'' is a little misleading since series include both derived and non derived lexemes.  Lexemes belong to the series on the basis of their form and meaning only.

Similarly, the representation of words that include interfixes such as \emph{tartelette} `little tart', \emph{gouttelette} `droplet', or \emph{vedettariat} `stardom' \citep{plenat2003.MMM3,plenat2005.decalage-ette} does not pose any difficulty.  These words are  full members of their respective families and series.  In these series, each of them is the nearest neighbor of its base \emph{tarte} `tart', \emph{goutte} `drop' and \emph{vedette} `star'.  The interfixes reinforce the formal integration of these lexemes in their series.

With respect to applications, the lexicon we propose adequately fulfills the main requirements for morphological knowledge in computational linguistics and information retrieval.  Morphological resources have several uses in these domains, such as prepositional phrase attachment disambiguation \citep{bourigault2007.HDR} or query expansion \citep{xu98.ACM,jing99.SIGIR,moreau2007.ECIR}.  The morphological relations used by a syntactic parser such as Syntex \citep{bourigault2007.HDR} associate nouns and verbs from the same family with strong morphological similarities.  Our lexicon will provide all these relations and even allow the users to select them with more precision.  In information retrieval, the retrieval performance can be improved by expanding the queries  by adding to them morphologically related words.  These words are all members of the morphological families of the words of the seed queries.  Besides, the morphological distance we propose can be used to tune the expansions more finely.  Our lexicon can also be used in the design of psycholinguistic experimental material.  The derived \emph{vs} non derived nature of the words can be determined from their derivational series and their morphological analogies.  Among the other features taken into account for the conception of experimental material, let us cite formal likeness andmembership in the same family.  All this information is explicitly available in the lexicon we propose.  Finally, let us stress that the relational organization of the lexicon does not pose any difficulty as proved by the number of the applications which use WordNet \citep{miller90.wordnet}.

\subsection{Related works}
\label{sec:related-works}

In this research, we adopt a global approach to the lexicon which differs from other efforts such as the MorTAL project aiming at creating a morphological database for French \citep{dal99.TALN,hathout2002.morphology-book}.  In this project, the database is made up by analyzing a selection of French affixes, one at a time, by means of the Dérif analyzer \citep{namer2005.HDR}.  By contrast, our objective is to create an entire lexicon at once.

Many works in the field of computational morphology aim to recover relations between lexical units.  All of them rely primarily on finding similarities between the word graphemic forms.  These relations are mainly prefixal or suffixal with two exceptions, \citep{yarowsky2000.ACL} and \citep{baroni2002.sigphon6}, who use string edit distances to estimate formal similarity.  As far as we know, all the others perform some sort of segmentation even when the goal is not to find morphemes, as in \citep{hathout2000.comlex} or \citep{neuvel2002.ACL}.  The model we propose differs from these approaches in that the graphemic similarities are determined solely on the basis of the sharing of graphemic features.  This is the main contribution of this paper.

This model is also related to approaches that combine graphemic and semantic cues in order to identify morphemes or morphological relations between words.  Usually, this semantic information is automatically acquired from corpora by means of various techniques such as latent semantic analysis \citep{schone2000.CoNLL}, mutual information \citep{baroni2002.sigphon6} or co-occurrence in $n$-word windows \citep{xu98.ACM,zweigenbaum2003.AIME}.  In the experiment presented here, semantic information is extracted from a machine readable dictionary and semantic similarity is calculated through random walks in a lexical graph.  The approach presented here can also be compared with \citep{hathout2002.LREC.wordnet,hathout2003.RIA}, where morphological knowledge is acquired by using semantic information extracted from dictionaries of synonyms and from WordNet.

\section{Morphological relatedness}
\label{sec:morph-relatedness}

\begin{figure}[h]
  \centerline{% Start of code
% \begin{tikzpicture}[anchor=mid,>=latex',join=bevel,]
\begin{tikzpicture}[>=latex',join=bevel,]
  \pgfsetlinewidth{1bp}
\begin{scope}
  \pgfsetstrokecolor{black}
\end{scope}
\begin{scope}
  \pgfsetstrokecolor{black}
\end{scope}
  \pgfsetcolor{black}
  % Edge: 8:e -> 3:w
  \draw [] (79bp,45bp) .. controls (147bp,45bp) and (163bp,45bp)  .. (230bp,45bp);
  % Edge: 8:e -> 4:w
  \draw [] (79bp,45bp) .. controls (148bp,45bp) and (162bp,9bp)  .. (230bp,9bp);
  % Edge: 8:e -> 5:w
  \draw [] (79bp,45bp) .. controls (148bp,45bp) and (162bp,81bp)  .. (230bp,81bp);
  % Edge: 7:e -> 3:w
  \draw [] (86bp,99bp) .. controls (155bp,99bp) and (162bp,45bp)  .. (230bp,45bp);
  % Edge: 7:e -> 4:w
  \draw [] (86bp,99bp) .. controls (162bp,99bp) and (155bp,9bp)  .. (230bp,9bp);
  % Edge: 7:e -> 5:w
  \draw [] (86bp,99bp) .. controls (151bp,99bp) and (166bp,81bp)  .. (230bp,81bp);
  % Edge: 7:e -> 6:w
  \draw [] (86bp,99bp) .. controls (151bp,99bp) and (166bp,117bp)  .. (230bp,117bp);
  % Edge: 7:e -> 1:w
  \draw [] (86bp,99bp) .. controls (155bp,99bp) and (144bp,189bp)  .. (212bp,189bp);
  % Edge: 7:e -> 2:w
  \draw [] (86bp,99bp) .. controls (140bp,99bp) and (141bp,153bp)  .. (194bp,153bp);
  % Edge: 9:e -> 3:w
  \draw [] (79bp,9bp) .. controls (148bp,9bp) and (162bp,45bp)  .. (230bp,45bp);
  % Edge: 9:e -> 4:w
  \draw [] (79bp,9bp) .. controls (147bp,9bp) and (163bp,9bp)  .. (230bp,9bp);
  % Edge: 10:e -> 6:w
  \draw [] (86bp,144bp) .. controls (152bp,144bp) and (165bp,117bp)  .. (230bp,117bp);
  % Edge: 10:e -> 2:w
  \draw [] (86bp,144bp) .. controls (135bp,144bp) and (146bp,153bp)  .. (194bp,153bp);
  % Edge: 11:e -> 1:w
  \draw [] (79bp,184bp) .. controls (139bp,184bp) and (153bp,189bp)  .. (212bp,189bp);
  % Edge: 11:e -> 2:w
  \draw [] (79bp,184bp) .. controls (132bp,184bp) and (142bp,153bp)  .. (194bp,153bp);
  % Node: 3
\begin{scope}
  \pgfsetstrokecolor{black}
  \draw (266bp,41bp) -- (266bp,49bp) -- (255bp,54bp) -- (241bp,54bp) -- (230bp,49bp) -- (230bp,41bp) -- (241bp,36bp) -- (255bp,36bp) -- cycle;
  \draw (248bp,45bp) node {\$or};
\end{scope}
  % Node: 4
\begin{scope}
  \pgfsetstrokecolor{black}
  \draw (266bp,5bp) -- (266bp,13bp) -- (255bp,18bp) -- (241bp,18bp) -- (230bp,13bp) -- (230bp,5bp) -- (241bp,0bp) -- (255bp,0bp) -- cycle;
  \draw (248bp,9bp) node {\$ori};
\end{scope}
  % Node: 5
\begin{scope}
  \pgfsetstrokecolor{black}
  \draw (266bp,77bp) -- (266bp,85bp) -- (255bp,90bp) -- (241bp,90bp) -- (230bp,85bp) -- (230bp,77bp) -- (241bp,72bp) -- (255bp,72bp) -- cycle;
  \draw (248bp,81bp) node {orient};
\end{scope}
  % Node: 6
\begin{scope}
  \pgfsetstrokecolor{black}
  \draw (266bp,113bp) -- (266bp,121bp) -- (255bp,126bp) -- (241bp,126bp) -- (230bp,121bp) -- (230bp,113bp) -- (241bp,108bp) -- (255bp,108bp) -- cycle;
  \draw (248bp,117bp) node {entati};
\end{scope}
  \pgfsetcolor{black}
  % Node: 1
\begin{scope}
  \pgfsetstrokecolor{black}
  \draw (284bp,198bp) -- (212bp,198bp) -- (212bp,180bp) -- (284bp,180bp) -- cycle;
  \draw (248bp,189bp) node {N.action\_X.de};
\end{scope}
  % Node: 2
\begin{scope}
  \pgfsetstrokecolor{black}
  \draw (302bp,162bp) -- (194bp,162bp) -- (194bp,144bp) -- (302bp,144bp) -- cycle;
  \draw (248bp,153bp) node {N.résultat\_X.de\_X.ce};
\end{scope}
  \pgfsetcolor{black}
  % Node: 7
\begin{scope}
  \pgfsetstrokecolor{black}
  \draw (43bp,99bp) ellipse (43bp and 9bp);
  \draw (43bp,99bp) node {N.orientation};
\end{scope}
  % Node: 8
\begin{scope}
  \pgfsetstrokecolor{black}
  \draw (43bp,45bp) ellipse (36bp and 9bp);
  \draw (43bp,45bp) node {V.orienter};
\end{scope}
  % Node: 9
\begin{scope}
  \pgfsetstrokecolor{black}
  \draw (43bp,9bp) ellipse (36bp and 9bp);
  \draw (43bp,9bp) node {A.original};
\end{scope}
  % Node: 10
\begin{scope}
  \pgfsetstrokecolor{black}
  \draw (43bp,144bp) ellipse (43bp and 9bp);
  \draw (43bp,144bp) node {N.fermentation};
\end{scope}
  % Node: 11
\begin{scope}
  \pgfsetstrokecolor{black}
  \draw (43bp,184bp) ellipse (36bp and 9bp);
  \draw (43bp,184bp) node {N.pointage};
\end{scope}
\end{tikzpicture}
% End of code
}
  \caption{\label{fig:schema-bigraphe}Excerpt of the bipartite graph which represents the lexicon.  Words are displayed in ovals, semantic features in rectangles and formal features in octagons.  The graph is symmetric.}
\end{figure}
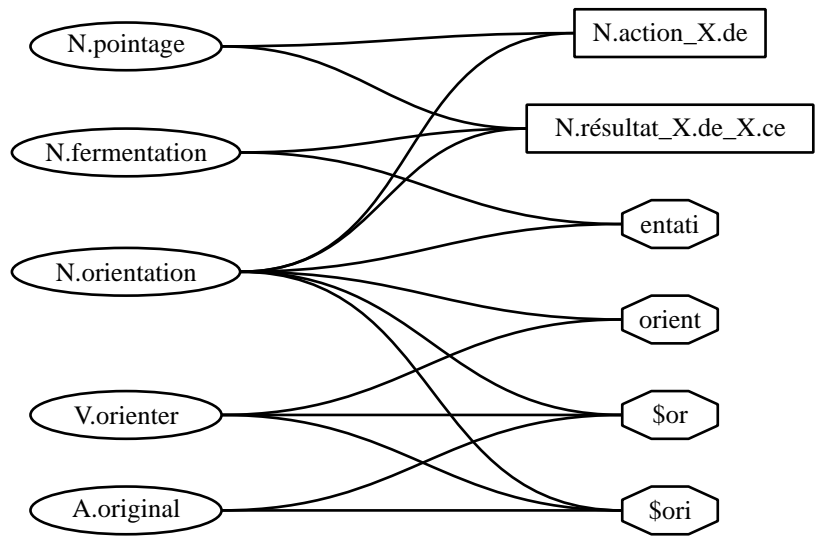

We assume here a minimalist definition of morphological relatedness: two words are morphologically related if they share phonological and semantic properties.  In the experiment, graphemic properties have been used instead of phonological ones because the TLFi does not provide the pronunciation of all the headwords.  The morphological relatedness is estimated by means of a bipartite graph like the one presented in figure~\ref{fig:schema-bigraphe}, with one subset of vertices representing lexemes and the other representing the formal and the semantic features of these lexemes.  Lexeme vertices are identified by the lemma and the grammatical category.

\subsection{Formal and semantic features}
\label{sec:formal-semantic-features}

The formal properties associated with a lexeme are the $n$-grams of letters that occur in its lemma.   The beginning and the end of the lemma are marked by the character \texttt{\$}.  We impose a minimum size on the $n$-grams ($n \ge 3$). For instance, the formal features associated with the French noun \emph{orientation} are the $n$-grams of figure~\ref{fig:traits-formels-orientation}, with $n$ ranging from $13$ down to $3$.  

\begin{figure}[h]
    \texttt{\$orientation\$}\\
    \texttt{\$orientation}~~ \texttt{orientation\$}\\
    \texttt{\$orientatio}~~ \texttt{orientation}~~ \texttt{rientation\$}\\
    \texttt{\$orientati}~~ \texttt{orientatio}~~ \texttt{rientation}~~ \texttt{ientation\$}\\
 $\cdots$\\
    \texttt{\$ori}~~ \texttt{orie}~~ \texttt{rien}~~ \texttt{ient}~~ \texttt{enta}~~ \texttt{ntat}~~ \texttt{tati}~~ \texttt{atio}~~ \texttt{tion}~~ \texttt{ion\$}\\
    \texttt{\$or}~~ \texttt{ori}~~ \texttt{rie}~~ \texttt{ien}~~ \texttt{ent}~~ \texttt{nta}~~ \texttt{tat}~~ \texttt{ati}~~ \texttt{tio}~~ \texttt{ion}~~ \texttt{on\$}
  \caption{\label{fig:traits-formels-orientation}Excerpt of the formal features associated with the noun \emph{orientation}.}
\end{figure}

\noindent{}Figure~\ref{fig:traits-formels-orientation} shows that the set of features associated with a given word is quite redundant.  An interesting property of this description is that it does not confer a special status to any of the individual $n$-grams which characterize the lexemes.  All $n$-grams play the same role and therefore none has the status of morpheme.  These features are only used to bring together the words that share the same sounds.

Alternatively, one could have used the $n$-grams that occur in the inflected forms of the lexemes as formal features.  Such an extended characterization is more faithful to word-based morphology and makes the inflectional allomorphies available at the derivational level.  However, we did not retain this option because inflectional endings reduce the homogeneity of the formal representations.  For instance, with a threshold $n \ge 3$, the verb \emph{malaxer} `knead' would become connected to all the words that contain \texttt{xie} (\emph{an\textbf{xie}ux}, \emph{le\textbf{xie}}, \emph{orthodo\textbf{xie}}, etc.) because of its inflected form \emph{mala\textbf{xie}z} (second person plural, imperfect indicative and present subjunctive).  In order to avoid giving too much importance to these very specific features, it is necessary to weight the contribution of each inflected form with an estimation of its frequency, computed for instance from a large text corpus.  A form like \emph{malaxiez} is likely to be very rare or even missing from most corpora.  In this way, the unwanted connections will be demoted or eliminated.

The semantic features associated with a lexeme are the $n$-grams of words that occur in its definitions.  The $n$-grams that contain punctuation marks, not counting apostrophes, are eliminated.  In other words, we only use $n$-grams of words that occur between two punctuation marks.   The words in the definitions are POS tagged and lemmatized. The tags are A for adjectives, N for nouns, R for adverbs, V for verbs and X for all other categories.  For instance, the semantic features induced by the definition \emph{Action d'orienter, de s'orienter\,; résultat de cette action} `act of directing, of finding one's way; result of this action' of the noun \emph{orientation} are presented in figure~\ref{fig:traits-semantiques-orientation}.   Notice that the semantic features are heavily redundant, just as the formal features are.

\begin{figure}[h]
  \texttt{N.action\_X.de\_V.orienter}~~ \texttt{N.action\_X.de}~~ \texttt{X.de\_V.orienter}\\ \texttt{N.action}~~ \texttt{X.de}~~ \texttt{V.orienter}~~ \texttt{X.de\_V.s'orienter}~~ \texttt{V.s'orienter}\\ \texttt{N.résultat\_X.de\_X.ce\_N.action}~~ \texttt{N.résultat\_X.de\_X.ce}~~ \texttt{X.de\_X.ce\_N.action}\\ \texttt{N.résultat\_X.de}~~ \texttt{X.de\_X.ce}~~ \texttt{X.ce\_N.action}~~ \texttt{N.résultat}~~ \texttt{X.ce}
  \caption{\label{fig:traits-semantiques-orientation}Semantic features induced by the definition \emph{Action d'orienter, de s'orienter\,; résultat de cette action} of the noun \emph{orientation}.}
\end{figure}

\noindent{}This is a very coarse semantic representation inspired from the repeated segments \citep{lebart98.kluwer}.  It offers several advantages:
\begin{enumerate}
\item being heavily redundant, it can capture various levels of similarity between the definitions;
\item it integrates information of a syntagmatic nature without a deep syntactic analysis of the definitions;
\item it slightly reduces the strong variations in the lexicographical treatment of the headwords, especially in the division into sub-senses and in the definitions.
\end{enumerate}

\subsection{Connecting the lexemes through their features}
\label{sec:connecting-lexemes}

The semantic and formal features are used in the same graph.  The bipartite graph is built up by connecting each headword to its semantic and formal features symmetrically.  For instance, the noun \emph{orientation} is connected with the formal features \texttt{\$or}, \texttt{\$ori}, \texttt{\$orie}, \texttt{\$orien}, etc. which are in turn connected with the words \emph{orienter}, \emph{orientable}, \emph{orientement} `orientation', \emph{orienteur} `orientator', etc.  Likewise, \emph{orientation} is connected with the semantic features \texttt{N.action X.de}, \texttt{N.résultat X.de X.ce N.action}, etc. which are themselves connected with the nouns \emph{orientement}, \emph{harmonisation}, \emph{pointage} `checking', etc.  The general schema is illustrated in figure~\ref{fig:schema-bigraphe}. It shows that the semantic and formal properties are used in the same manner.  This representation corresponds precisely to the Network Model of \cite{bybee88.lexical-organization,bybee95.regular-morphology}.

Actually, the bipartite structure is not essential.  All we need is to be able to compute a morphological distance between the words.  We use a bipartite graph mainly because it allows us to spread an activation simultaneously into the formal and the semantic subparts of the graph.  The graph is also interesting because it contains representations of properties that are useful for morphological studies.  They could for instance be used to describe the semantics of the \emph{-able} suffixation or to find the characteristic endings of boat names in French (\emph{voilier}, \emph{pétrolier}, \emph{bananier}, \emph{thonier}, \emph{sardinier}\ldots;  \emph{patrouilleur}, \emph{torpilleur}, \emph{caboteur}, \emph{dériveur}, \emph{dragueur}\ldots).

\subsection{Estimating the morphological similarity between words }
\label{sec:estim-morph-simil}

The morphological similarity between a word and its neighbors is estimated by simulating the spreading of an activation initiated at the vertex that represents that word. Since the graph is bipartite, the activation has to be propagated an even number of times.  The graph being heavily redundant, two steps of propagation are sufficient to obtain the intended proximity estimations.

For instance, if we want to determine what the closest neighbors of \emph{orientation} are, we initiate an activation at the vertex that represents \emph{orientation}.  Then, this activation is uniformly spread toward the formal and semantic features of \emph{orientation}.  In the next step, the activation located on the feature vertices is spread toward the lexeme vertices.  The greater the number of features shared by a lexeme with \emph{orientation} and the more specific these features are, the stronger the activation it receives.  The assumption is that the strength of the activation is an estimation of the degree of morphological relatedness.

Technically, the spreading is simulated as a random walk in the graph \citep{gaume2002.JETAI,gaume2005.hierarchy-lexical-organization,muller2006.textgraphs}.  It is classically computed as a multiplication of the stochastic adjacency matrix of the graph. More precisely, let $G=(V,E)$ be a graph consisting of a set of vertices $V=\{v_1,\ldots,v_n\}$ and a set of edges $E \subset V\times V$.  Let $A$ be the adjacency matrix of $G$, that is a $n \times n$ matrix such that $A_{ij}=1$ if $(v_i,v_j)\in E$ and $A_{ij}=0$ if $(v_i,v_j)\not\in E$.    We normalize the rows of $A$ in order to get a stochastic matrix $M$:
\[\forall i \in [1,n], \forall j \in [1,n], M_{ij}=\dfrac{A_{ij}}{\sum\limits_{k=1}^n A_{ik}}\]

\noindent{}Then  $(M^n)_{ij}$ is the probability of reaching vertex $v_j$ from the vertex $v_i$ through a walk of $n$ steps.  This probability can also be regarded as an activation level of node $v_j$ following an $n$-step spreading initiated at node $v_i$.

In the experiment presented in this paper, one half of the activation is spread toward the semantic features and the other half toward the formal features.  The edges of the bipartite graph can be divided into three parts $E=J \cup K \cup L$ where $J$ contains the edges that connect a headword to a formal feature, $K$ the edges that connect a headword to a semantic feature and $L$ the edges that connect a formal or semantic feature to a headword.  The actual values of $M$ are defined as follows:
\[\begin{array}{lcl}
  \mbox{if $e_{ij} = (v_i,v_j) \in J$, } & M_{ij}= &\left\{%
    \begin{array}{ll}
      0.5 \, \dfrac{A_{ij}}{\sum\limits_{e_{ih} \in J} A_{ih}} & \mbox{if  $v_i$ is connected to a semantic feature}\\[4ex]
      \dfrac{A_{ij}}{\sum\limits_{e_{ik} \in J} A_{ik}} & \mbox{otherwise}
    \end{array}%
  \right.\\[7ex]
\mbox{if $e_{ik} = (v_i,v_k) \in K$, }  & M_{ik}= &\left\{%
  \begin{array}{ll}
    0.5 \, \dfrac{A_{ik}}{\sum\limits_{e_{ih} \in K} A_{ih}}  & \mbox{if $v_i$ is connected to a formal feature} \\[4ex]
    \dfrac{A_{ik}}{\sum\limits_{e_{ih} \in K} A_{ih}}  & \mbox{otherwise}
  \end{array}%
\right.\\[7ex]
\mbox{if $e_{il} = (v_i,v_l) \in L$, } & M_{il}= &\dfrac{A_{il}}{\sum\limits_{e_{ih} \in L} A_{ih}}
\end{array}
\]

\subsection{Morphological neighbors}
\label{sec:morph-neighb}

The graph used in the experiment was built from the headwords and the definitions of the TLFi.  We only removed the definitions of non standard uses (old, slang, regionalism, etc.).  The extraction and cleaning-up of the definitions were carried out in collaboration with Bruno Gaume and Philippe Muller.  The bipartite graph was created from  225~529 definitions describing 75~024 headwords (lexemes).  They induced about 9 million features, 90\% of them being associated with only one headword.  These features were removed because they do not contribute to  the connections of different headwords.   Table~\ref{tab:nb-traits} shows that this reduction is stronger for the semantic features (93\%) than it is for the formal ones (69\%).  Indeed, semantic descriptions show greater variability than formal ones.
\begin{table}[h]
  \centerline{
        \begin{tabular}{lrrr}
          \hline
          features & complete & reduced & hapax\\
          \hline
          formal   & 1~306~497 & 400~915 & 69\%\\
          semantic & 7~650~490 & 548~641 & 93\%\\
          \hline
          total    & 8~956~987 & 949~556 & 90\%\\
          \hline
        \end{tabular}
  }
  \caption{\label{tab:nb-traits}Numbers of semantic and formal features.}
\end{table}

The use of the graph is illustrated in figure~\ref{fig:voisins-fructifier}.  It shows the 40 nearest neighbors of the verb \emph{fructifier} `bear fruit' for three propagation configurations.  The first row (\emph{form}) presents the neighbors of \emph{fructifier} in a graph that only contains formal features.  It shows that the members of the morphological family tend to appear as the closest neighbors and that the members of the derivational series (i.e. the verbs ending in \emph{-ifier}) are more distant.  The members in the second row (\emph{sem}) have been computed in a graph that only contains the semantic features and the ones in the third row (\emph{form + sem}) in the full graph.

\begin{figure}[h]
  \begin{tabular}{lp{.825\textwidth}}
   form &   \textbf{V.fructifier N.fructification A.fructificateur A.fructifiant A.fructifère V.sanctifier V.rectifier} \emph{A.rectifier V.fructidoriser N.fructidorien N.fructidor} \textbf{N.fructuosité R.fructueusement A.fructueux} \emph{N.rectifieur A.obstructif A.instructif A.destructif A.constructif} \textbf{N.infructuosité R.infructueusement A.infructueux V.transsubstantifier V.substantifier V.stratifier V.schistifier V.savantifier V.refortifier V.ratifier V.quantifier V.présentifier V.pontifier V.plastifier V.notifier V.nettifier V.mystifier V.mortifier V.justifier V.idiotifier V.identifier} \\
&\\

  sem &  \textbf{V.fructifier} \emph{V.trouver N.missionnaire N.mission A.missionnaire N.saisie N.police N.hangar N.dîme N.ban V.affruiter N.melon N.saisonnement N.azédarach A.fruitier A.bifère V.saisonner N.roman N.troubadour V.contaminer N.conductibilité N.alevinage V.profiter} \textbf{A.fructifiant} \emph{N.pouvoir V.agir N.opération V.placer N.rentabilité N.jouissance N.avocat N.report} \textbf{A.fructueux} \emph{V.tourner V.chiper N.économat N.visa N.société N.réserve N.récréance} \\
&\\
  form + sem &  \textbf{V.fructifier A.fructifiant N.fructification A.fructificateur} \emph{V.trouver} \textbf{A.fructifère V.rectifier V.sanctifier} \emph{A.rectifier V.fructidoriser N.fructidor N.fructidorien N.missionnaire N.mission A.missionnaire} \textbf{A.fructueux R.fructueusement N.fructuosité} \emph{N.rectifieur N.saisie N.police N.hangar N.dîme N.ban A.fruitier V.affruiter A.instructif A.obstructif A.destructif A.constructif N.conductibilité V.saisonner N.melon N.saisonnement N.azédarach A.bifère V.contaminer N.roman N.troubadour N.alevinage}
  \end{tabular}
  \caption{\label{fig:voisins-fructifier}The 40 nearest neighbors of the verb \emph{fructifier} when the activation is spread only toward the formal features in the first row, only toward the semantic ones in the second row and toward both the semantic and formal features in the third.  Words that belong to the family or series of \emph{fructifier} are in boldface; the others are in italic.}
\end{figure}

\noindent{}The first two rows show clearly that formal features are the more predictive ones while semantic features are the less reliable ones. These examples provide an insight into some of the features of the morphological families and the derivational series that could be used in order to separate them: families are small sets; series are larger sets; families have a strong semantic and formal cohesion; members of a series have looser semantic and formal connections.  The last two features explain why the members of the morphological families tend to show up before the members of the derivational series.  The examples also show that the morphological similarity is not selective enough and that the list of neighbors cannot be used as is.  We need to further filter them and we propose to do so with formal analogy.

\section{Analogy}
\label{sec:analogy}

\subsection{Familial and serial analogies}
\label{sec:famil-seri-anal}

The members of the series and families are massively involved in analogies which structure the lexicon.  For instance, \emph{fructifier} and \emph{fructification} which belong to the same family form analogies with large numbers of pairs of members of other families (\emph{rectifier} `correct', \emph{rectification}), (\emph{certifier} `assure', \emph{certification} `attestation'), (\emph{plastifier} `coat with plastic', \emph{plastification} `lamination of document'), (\emph{sanctifier}, \emph{sanctification}), (\emph{vitrifier}, \emph{vitrification}), etc.  Besides, the first elements of each of these pairs belong to the series of \emph{fructifier} and the second ones to the series of \emph{fructification}.  In a dual manner, \emph{fructifier} and \emph{sanctifier} which belong to the same series form analogies with the members of other series (\emph{fructificateur} `fructifier', \emph{sanctificateur} `sanctifier'), (\emph{fructification}, \emph{sanctification}) or (\emph{fructifiant} `fructifying', \emph{sanctifiant} `sanctifying').  These pairs are respectively made of members of the families of \emph{fructifier} and \emph{sanctifier}.

\begin{figure}[h]
% dot2tex --tikzedgelabels analogie-fructifier1.dot > analogie-fructifier1.tex;pdflatex analogie-fructifier1.tex
% dot2tex --tikzedgelabels analogie-fructifier2.dot > analogie-fructifier2.tex;pdflatex analogie-fructifier2.tex
% supprimer les entêtes et la fin des fichiers latex
\centerline{\mbox{}\hfill% Start of code
% \begin{tikzpicture}[anchor=mid,>=latex',join=bevel,]
\begin{tikzpicture}[>=latex',join=bevel,]
  \pgfsetlinewidth{1bp}
\begin{scope}
  \pgfsetstrokecolor{black}
\end{scope}
\begin{scope}
  \pgfsetstrokecolor{black}
\end{scope}
  \pgfsetcolor{black}
  % Edge: 1:e -> 3:w
  \draw [below,sloped] (63bp,72bp) .. controls (86bp,72bp) and (92bp,72bp)  .. node {family} (114bp,72bp);
  % Edge: 2:e -> 4:w
  \draw [above,sloped] (61bp,12bp) .. controls (86bp,12bp) and (93bp,12bp)  .. node {family} (117bp,12bp);
  % Edge: 1 -> 2
  \draw [below,sloped] (32bp,24bp) .. controls (32bp,34bp) and (32bp,49bp)  .. node {series} (32bp,60bp);
  % Node: 1
\begin{scope}
  \pgfsetstrokecolor{black}
  \draw (32bp,72bp) node {fructifier};
\end{scope}
  % Node: 2
\begin{scope}
  \pgfsetstrokecolor{black}
  \draw (32bp,12bp) node {rectifier};
\end{scope}
  \pgfsetcolor{black}
  % Edge: 3 -> 4
  \draw [above,sloped] (157bp,24bp) .. controls (157bp,34bp) and (157bp,49bp)  .. node {series} (157bp,60bp);
  % Node: 3
\begin{scope}
  \pgfsetstrokecolor{black}
  \draw (157bp,72bp) node {fructification};
\end{scope}
  % Node: 4
\begin{scope}
  \pgfsetstrokecolor{black}
  \draw (157bp,12bp) node {rectification};
\end{scope}
\end{tikzpicture}
% End of code
\hfill% Start of code
% \begin{tikzpicture}[anchor=mid,>=latex',join=bevel,]
\begin{tikzpicture}[>=latex',join=bevel,]
  \pgfsetlinewidth{1bp}
\begin{scope}
  \pgfsetstrokecolor{black}
\end{scope}
\begin{scope}
  \pgfsetstrokecolor{black}
\end{scope}
  \pgfsetcolor{black}
  % Edge: 1:e -> 3:w
  \draw [->,below,sloped] (63bp,72bp) .. controls (88bp,72bp) and (98bp,72bp)  .. node {neighbor} (128bp,72bp);
  % Edge: 2:e -> 4:w
  \draw [->,above,sloped] (61bp,12bp) .. controls (89bp,12bp) and (98bp,12bp)  .. node {neighbor} (131bp,12bp);
  % Edge: 1 -> 2
  \draw [<-,below,sloped] (32bp,24bp) .. controls (32bp,43bp) and (32bp,52bp)  .. node {neighbor} (32bp,60bp);
  % Node: 1
\begin{scope}
  \pgfsetstrokecolor{black}
  \draw (32bp,72bp) node {fructifier};
\end{scope}
  % Node: 2
\begin{scope}
  \pgfsetstrokecolor{black}
  \draw (32bp,12bp) node {rectifier};
\end{scope}
  \pgfsetcolor{black}
  % Edge: 3 -> 4
  \draw [<-,above,sloped] (171bp,24bp) .. controls (171bp,43bp) and (171bp,52bp)  .. node {neighbor} (171bp,60bp);
  % Node: 3
\begin{scope}
  \pgfsetstrokecolor{black}
  \draw (171bp,72bp) node {fructification};
\end{scope}
  % Node: 4
\begin{scope}
  \pgfsetstrokecolor{black}
  \draw (171bp,12bp) node {rectification};
\end{scope}
\end{tikzpicture}
% End of code
\hfill\mbox{}}
  \caption{\label{fig:analogie-fructifier}Morphological relations and neighborhood relations between the members of the \emph{fructifier}:\emph{fructification} :: \emph{rectifier}:\emph{rectification} analogy.}
\end{figure}
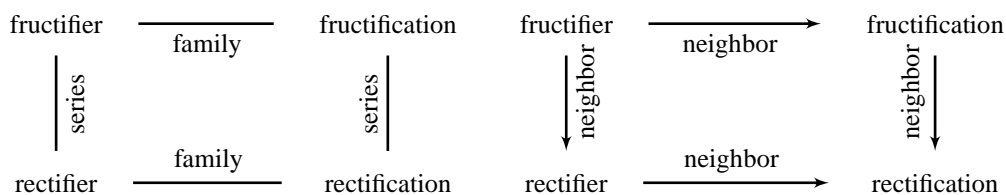

Formal analogies can be used in order to filter the morphological neighbors of a word.  Actually, we are interested in analogies such as \emph{fructifier}:\emph{fructification} :: \emph{rectifier}:\emph{rectification}.  Since \emph{fructification} belongs to the family of \emph{fructifier} and \emph{rectifier} to its series, both are morphological neighbors of \emph{fructifier}.  Similarly, \emph{rectification} belongs to the series of \emph{fructification} and to the family of \emph{rectifier}.  Therefore, it is a morphological neighbor of both \emph{fructification} and \emph{rectifier}.  These relations are illustrated in figure~\ref{fig:analogie-fructifier}.  Conversely, if we consider that the morphological neighbors of a word are likely to be morphologically related to that word, then we can use them to look for quadruplets that could form analogies.  These quadruplets could be found as follows:

For a given word $a$,\\
\indent{}\indent{}look for two of its neighbors $b$ and $c$, then\\
\indent{}\indent{}for every $d$ that is a neighbor of both $b$ and $c$,\\
\indent{}\indent{}\indent{}the quadruplet $a:b::c:d$ is likely to be an analogy.

\noindent{}More generally,  if $b$ is a correct morphological neighbor of $a$, then it is either a member of the family of $a$ or a member of its series.  Therefore, there exists another neighbor $c$ of $a$ ($c$ belongs to the family of $a$ if $b$ belongs to the series of $a$ or vice versa) such that there exists a neighbor $d$ of $b$ and of $c$ such that $a:b::c:d$. We then have only two configurations:
\begin{enumerate}
\item if $b \in F_a$, then $\exists c \in S_a, \exists d \in S_b \cap F_c, a:b::c:d$
\item if $b \in S_a$, then $\exists c \in F_a, \exists d \in F_b \cap S_c, a:b::c:d$
\end{enumerate}
where $F_x$ is the morphological family of $x$ and $S_x$ the derivational series of $x$.

\subsection{Formal analogy}
\label{sec:formal-analogy}

A formal or graphemic analogy is a relation $a:b::c:d$ that holds between four strings such that the graphemic differences between $a$ and $b$ are the same as the ones between $c$ and $d$.  This is the case for \emph{fructifier}:\emph{fructification} :: \emph{rectifier}:\emph{rectification} (see figure~\ref{fig:exemple-analogie-formelle}).  Naturally, more than one difference can appear in the pair as with the four Arabic words \emph{kataba}:\emph{maktoubon} :: \emph{fa3ala}:\emph{maf3oulon} which respectively are transcriptions of the verb `write', the noun `document', the verb `do' and the noun `effect.'\footnote{%
  This example is adapted from examples in \cite{lepage98.COLING,lepage2003.HDR}.%
} The differences between the first two words and between the two last ones can be described as in figure~\ref{fig:exemple-analogie-formelle}.  They are identical for the two pairs of words.  This example shows that even analogies in a templatic language like Arabic can be checked in this way.
\begin{figure}[h]
  \centerline{\mbox{}\hfill\texttt{
    \begin{tabular}{l|l|}
      \cline{2-2}
      fructifi & er  \\
      fructif & cation \\
      \cline{2-2}
      \multicolumn{2}{l}{} \\
      \cline{2-2}
      rectifi & er \\
      rectifi & cation \\
      \cline{2-2}
    \end{tabular}\hfill
    \begin{tabular}{|l|l|l|l|l|l|l|}
      \cline{1-1}\cline{3-3}\cline{5-5}\cline{7-7}
      $\epsilon$ & k & a          & t & a  & b & a  \\
      ma         & k & $\epsilon$ & t & ou & b & on \\
      \cline{1-1}\cline{3-3}\cline{5-5}\cline{7-7}
      \multicolumn{6}{l}{} \\
      \cline{1-1}\cline{3-3}\cline{5-5}\cline{7-7}
      $\epsilon$ & f & a          & 3 & a  & l & a  \\
      ma         & f & $\epsilon$ & 3 & ou & l & on \\
      \cline{1-1}\cline{3-3}\cline{5-5}\cline{7-7}
    \end{tabular}
  }\hfill\mbox{}}
  \caption{\label{fig:exemple-analogie-formelle}Formal analogies  \emph{fructifier}:\emph{fructification} :: \emph{rectifier}:\emph{rectification} and \emph{kataba}:\emph{maktoubon} :: \emph{fa3ala}:\emph{maf3oulon}. The differences are located in the frame boxes. $\epsilon$ represents the empty string.}
\end{figure}

More generally, formal analogies can be defined in terms of factorization \citep{stroppa2005.CoNLL}.  Let $L$ be an alphabet and $a \in L^{\star}$ a string over $L$.  A factorization of $a$ is a sequence $f=(f_1, \cdots, f_n) \in {L^{\star}}^{n}$ such that $a=f_1 \oplus \cdots \oplus f_n$ where $\oplus$ denotes concatenation.  For instance, (\texttt{ma}, \texttt{k}, $\epsilon$, \texttt{t}, \texttt{ou}, \texttt{b}, \texttt{on}) is a factorization of length $7$ of \texttt{maktoubon}.  Morphological analogies can be defined as follows.  Let $(a,b,c,d) \in {L^{\star}}^4$ be four strings. $a:b::c:d$ is a formal analogy iff there exists $n \in \mathbb{N}$ and four factorizations of length $n$ of the four strings $(f(a),f(b),f(c),f(d)) \in ({L^{\star}}^n)^4$ such that, $\forall i \in [1,n], (f_i(b),f_i(c)) \in \{(f_i(a),f_i(d)),(f_i(d),f_i(a))\}$.  For the analogy \emph{kataba}:\emph{maktoubon} :: \emph{fa3ala}:\emph{maf3oulon}, the property holds for $n=7$ (see figure~\ref{fig:exemple-analogie-formelle}).

\subsection{Implementation}
\label{sec:implementation}

Formal analogies are checked at the graphemic level. The differences between the first and second pairs of strings are calculated from the sequence of string edit operations that transform the first form of each pair into the second one. Both sequences must minimize Levenshtein edit distance (i.e. have the least cost). Each sequence corresponds to a path in the edit lattices of the pair of words.  The lattices are represented by a matrix computed using the standard string edit algorithm \citep{jurafsky2000.speech-language-processing}. The path which describes the sequence of string edit operations starts at the last cell of the matrix and climbs to the first one.  It is made up as follows: for each cell, select the neighboring one with the least cost ; in case of equal costs, prefer the cell to the left (insertion), then the one upward (deletion) and otherwise the one in the upper left diagonal direction (substitution).  Figure~\ref{fig:string-edit-path} presents the path that is selected in the string edit matrix of \emph{fructueux} `fruitful' and \emph{infructueusement} `fruitlessly' and figure~\ref{fig:sequence-operations-fructueux-infructueusement}, the sequence of edit operations for this pair.

\begin{figure}[h]
    \centerline{\begin{tabular}{c|c@{}c@{}c@{}c@{}c@{}c@{}c@{}c@{}c@{}c@{}c@{}c@{}c@{}c@{}c@{}c@{}c@{}c@{}c@{}c@{}c@{}c@{}c@{}c@{}c@{}c@{}c@{}c@{}c@{}c@{}c@{}c@{}c}
      & $\diamond$ && i && n && f && r && u && c && t && u && e && u && s && e  && m  && e  && n  && t \\
      \hline
      $\diamond$  & 0 & $\leftarrow$ & 1 & $\leftarrow$ & 2 & \raisebox{-1.5ex}{\scriptsize$\nwarrow$} & 3 && 4 && 5 && 6 && 7 && 8 && 9 && 10 && 11 && 12 && 13 && 14 && 15 && 16 \\
      f & 1 && 1 && 2 && 2 &\raisebox{-1.5ex}{\scriptsize$\nwarrow$} & 3 && 4 && 5 && 6 && 7 && 8 && 9 && 10 && 11 && 12 && 13 && 14 && 15 \\
      r & 2 && 2 && 2 && 3 && 2 &\raisebox{-1.5ex}{\scriptsize$\nwarrow$} & 3 && 4 && 5 && 6 && 7 && 8 && 9 && 10 && 11 && 12 && 13 && 14 \\
      u & 3 && 3 && 3 && 3 && 3 && 2 &\raisebox{-1.5ex}{\scriptsize$\nwarrow$} & 3 && 4 && 5 && 6 && 7 && 8 && 9 && 10 && 11 && 12 && 13 \\
      c & 4 && 4 && 4 && 4 && 4 && 3 && 2 &\raisebox{-1.5ex}{\scriptsize$\nwarrow$} & 3 && 4 && 5 && 6 && 7 && 8 && 9 && 10 && 11 && 12 \\
      t & 5 && 5 && 5 && 5 && 5 && 4 && 3 && 2 &\raisebox{-1.5ex}{\scriptsize$\nwarrow$} & 3 && 4 && 5 && 6 && 7 && 8 && 9 && 10 && 11 \\
      u & 6 && 6 && 6 && 6 && 6 && 5 && 4 && 3 && 2 &\raisebox{-1.5ex}{\scriptsize$\nwarrow$} & 3 && 4 && 5 && 6 && 7 && 8 && 9 && 10 \\
      e & 7 && 7 && 7 && 7 && 7 && 6 && 5 && 4 && 3 && 2 &\raisebox{-1.5ex}{\scriptsize$\nwarrow$} & 3 && 4 && 5 && 6 && 7 && 8 && 9 \\
      u & 8 && 8 && 8 && 8 && 8 && 7 && 6 && 5 && 4 && 3 && 2 & \raisebox{-1.5ex}{\scriptsize$\nwarrow$} & 3 && 4 && 5 && 6 && 7 && 8 \\
      x & 9 && 9 && 9 && 9 && 9 && 8 && 7 && 6 && 5 && 4 && 3 && 3 & $\leftarrow$ & 4 & $\leftarrow$ & 5 & $\leftarrow$ & 6 & $\leftarrow$ & 7 & $\leftarrow$ & 8
    \end{tabular}}
  \caption{\label{fig:string-edit-path}Least cost path describing a sequence of string edit operations that transforms \emph{fructueux} into \emph{infructueusement}. $\diamond$ represents the beginning of the string.  Cell $(i,j)$ in the matrix indicates the Levenshtein distance between the substring consisting of the first $i$ characters of \emph{fructueux} and the one consisting of the first $j$ characters of \emph{infructueusement}.}
\end{figure}

\begin{figure}[h]
  \centerline{
    \begin{tabular}{ccccccccccccccccc}
      I & I & M & M & M & M & M & M & M & M & S & I & I & I & I & I \\
      \hline
      $\epsilon$ & $\epsilon$ & \texttt{f} & \texttt{r} & \texttt{u} & \texttt{c} & \texttt{t} & \texttt{u} & \texttt{e} & \texttt{u} & \texttt{x} & $\epsilon$ & $\epsilon$ & $\epsilon$ & $\epsilon$ & $\epsilon$ \\
      \texttt{i} & \texttt{n} & \texttt{f} & \texttt{r} & \texttt{u} & \texttt{c} & \texttt{t} & \texttt{u} & \texttt{e} & \texttt{u} & \texttt{s} & \texttt{e} & \texttt{m} & \texttt{e} & \texttt{n} & \texttt{t}
    \end{tabular}
  }
  \caption{\label{fig:sequence-operations-fructueux-infructueusement}Sequence of edit operations that transform \emph{fructueux} into \emph{infructueusement}.  The types of the operations are indicated on the first row: D for deletion, I for insertion, M for matching and S for a substitution by a different character.}
\end{figure}

Sequences of edit operations can be simplified by merging the series of identical character matchings.  The sequence in figure~\ref{fig:sequence-operations-fructueux-infructueusement} then becomes (\ref{enum:string-edit-1}).   This simplified sequence is identical to the one for the pair \emph{soucieux}:\emph{insoucieusement} `worried':`unworriedly' except for the matching operation (\ref{enum:string-edit-2}).

\enumsentence{\label{enum:string-edit-1}
 ((I,$\epsilon$,\texttt{i}), (I,$\epsilon$,\texttt{n}), (M,\texttt{fructueu},\texttt{fructueu}), (S,\texttt{x},\texttt{s}), (I,$\epsilon$,\texttt{e}), (I,$\epsilon$,\texttt{m}), (I,$\epsilon$,\texttt{e}), (I,$\epsilon$,\texttt{n}), (I,$\epsilon$,\texttt{t}))}
\enumsentence{\label{enum:string-edit-2}
  ((I,$\epsilon$,\texttt{i}), (I,$\epsilon$,\texttt{n}), (M,\texttt{soucieu},\texttt{soucieu}), (S,\texttt{x},\texttt{s}), (I,$\epsilon$,\texttt{e}), (I,$\epsilon$,\texttt{m}), (I,$\epsilon$,\texttt{e}), (I,$\epsilon$,\texttt{n}), (I,$\epsilon$,\texttt{t}))}

\noindent{}The two sequences can be made identical if the matching sub-strings are not specified (i.e. replaced by a wildcard character @).  The resulting sequence can then be assigned to both pairs  as their edit signatures ($\sigma$). The formal analogy \emph{fructueux}:\emph{infructueusement} :: \emph{soucieux}:\emph{insoucieusement} can be stated in terms of identity of the edit signatures of the two pairs (\ref{enum:signature}).

\enumsentence{\label{enum:signature}$\sigma(\texttt{fructueux},\texttt{infructueusement}) =$\\
$\sigma(\texttt{soucieux},\texttt{insoucieusement}) = $\\
((I,$\epsilon$,\texttt{i}), (I,$\epsilon$,\texttt{n}), (M,@,@), (S,\texttt{x},\texttt{s}), (I,$\epsilon$,\texttt{e}), (I,$\epsilon$,\texttt{m}), (I,$\epsilon$,\texttt{e}), (I,$\epsilon$,\texttt{n}), (I,$\epsilon$,\texttt{t}))
}

\noindent{}More generally, four strings  $(a,b,c,d) \in {L^{\star}}^4$ form a formal analogy $a:b::c:d$ iff $\sigma(a,b)=\sigma(c,d)$.

\section{First results}
\label{sec:first-results}

This is work in progress and we only have preliminary results.  We have computed the 100 nearest neighbors of the headwords of the TLFi, then collected the formal analogies for 22 headwords belonging to 4 morphological families and checked them manually.  An analogy $a:b::c:d$ is accepted as correct if:
\begin{itemize}
\item $b$ belongs to the family of $a$, $c$ belongs to the series of $a$, $d$ belongs to series of $b$ and to the family of $c$, or
\item $b$ belongs to the series of $a$, $c$ belongs to the family of $a$, $d$ belongs to family of $b$ and to the series of $c$.
\end{itemize}
We present some examples of correct analogies in (\ref{ex:analogies-correctes}) and erroneous ones in (\ref{ex:analogies-incorrectes}).  We can see that the collected analogies involve words that are derived one from the other (\ref{ex:analogies-correctes-1}), words that are derived from a common base (\ref{ex:analogies-correctes-2}) and words connected through a sequence of derivations (\ref{ex:analogies-correctes-3}).

\eenumsentence{\label{ex:analogies-correctes}
\item\label{ex:analogies-correctes-1} N.fructification:N.identification :: V.fructifier:V.identifier
\item\label{ex:analogies-correctes-2} A.fructifiant:A.fructificateur :: A.glorifiant :: A.glorificateur
\item\label{ex:analogies-correctes-3} A.fructueux:A.affectueux :: N.infructuosité:N.inaffectuosité
\item A.frugivore:A.végétivore :: R.frugalement:R.végétalement
\item A.fruitarien:A.végétarien :: N.fruitarisme:N.végétarisme
\item A.fruitier:A.laitier :: N.fruiterie:N.laiterie
\item R.fructueusement:R.affectueusement :: N.fructuosité:N.affectuosité	
}
\eenumsentence{\label{ex:analogies-incorrectes}
\item A.fruité:N.fruste :: A.truité:N.truste
\item N.fruit:N.frumentaire :: A.instruit:A.instrumentaire	
\item N.fruiterie:N.friterie :: V.effruiter:V.effriter
}

We have tested three configurations (see §~\ref{sec:morph-neighb}). In the first, we have used neighbors from the graph that contains the formal features only, in the second, the semantic features only, and in the third, both the formal and the semantic features.  The results are summarized in table~\ref{tab:eval-analogies}.  Their quality is quite satisfactory.  We observe that the number of analogies depends on the configuration of propagation.  The use of the semantic features improves the precision but reduces the total number of analogies that are collected.  The best trade-off is a simultaneous propagation toward the semantic and the formal features.

\begin{table}[h]
  \centerline{
    \begin{tabular}{|l|r|r|r|}
      \hline
      configuration    & analogies & correct & errors \\
      \hline\hline
      formal           &    169 &      163 & 3.6\%\\
      semantics        &      5 &        5 & 0.0\%\\
      sem $+$ form     &    130 &      128 & 1.5\%\\
      \hline
    \end{tabular}
  }
  \caption{\label{tab:eval-analogies}Number of the analogies collected for a sample of 22 headwords and error rate.}
\end{table}

The performance of the method strongly depends on the length of the headwords because the method mainly relies on formal similarity and because formal similarity is stronger for long words.   Table~\ref{tab:eval-analogies-longueur} show this correlation clearly.   It presents the number of analogies and the error rate of 13 samples of 5 words, selected randomly.  The analogies have been collected from neighborhoods in the full graph. The words in each group are of the same length.  Lengths range from 4 to 16 letters. We can see that the analogies collected for words of 10 letters or more are all correct.
\begin{table}[h]
  \centerline{
    \begin{tabular}{|r|r|r|r|}
      \hline
      length & analogies & correct & errors \\
      \hline\hline
        4          & 29 &  14 & 51.7\% \\
        5          & 22 &  14 & 36.4\% \\
        6          &  8 &   7 & 12.5\% \\
        7          & 10 &   8 & 20.0\% \\
        8          & 55 &  54 &  1.8\% \\
        9          & 29 &  27 &  6.9\% \\
        10         & 30 &  30 &  0.0\% \\
        11         & 32 &  32 &  0.0\% \\
        12         & 19 &  19 &  0.0\% \\
        13         & 11 &  11 &  0.0\% \\
        14         & 35 &  35 &  0.0\% \\
        15         & 63 &  63 &  0.0\% \\
        16         & 39 &  39 &  0.0\% \\
      \hline
    \end{tabular}
  }
  \caption{\label{tab:eval-analogies-longueur}Numbers of analogies and error rates for headwords of length 4 to 16.}
\end{table}

\section{Conclusion and directions for further research}
\label{sec:future-prospects}

We have presented a computational model that makes the morphological structure of the lexicon emerge from the formal and semantic properties of the words it contains.  The model is radically word-based.  It integrates the semantic and formal properties of the words in a uniform manner and represents them in a bipartite graph.  Random walks are used to simulate the spreading of activations in the lexical network.  The level of activation obtained after the propagation indicates the lexical relatedness of the words.  The members of the morphological family and the derivational series of a word are then identified among its lexical neighbors by means of formal analogies.

Let us stress that this method is promising because it is mainly computational.  Almost no theoretical assumptions have been made. The method primarily exploits the memory and the computing power of the processors.  Another interesting feature is that the formal and semantic properties of the words are represented separately.  Therefore, the method deals with so-called parasynthetic derivatives like any other lexemes.

The next steps of this research are to create an initial network with only long words and then use a bootstrap method.  One important task that remains to be done is to separate the members of the families from the ones of the series.  We also intend to conduct a similar experiment on the English lexicon and to evaluate our results in a more classical manner by using the CELEX database \citep{baayen95.CELEX} as gold standard. The evaluation should also be done with respect to well-known systems like \emph{Linguistica} \citep{goldsmith2001.CL} or the morphological analyzer of \cite{bernhard2006.FinTAL}.

\section*{Acknowledgments}
\label{sec:acknoledgments}

I would like to thank the ATILF laboratory and Jean-Marie Pierrel for making available to me the TLFi. I am in debt to Bruno Gaume and Philippe Muller for the many discussions and exchanges we have had on the cleaning-up of the TFLi and its exploitation through random walks.  I am also grateful to Gilles Boyé, Olivier Haute-Cœur, Ludovic Tanguy, Jesse Tseng and audiences at University of Toulouse, University of Bordeaux, the Textgraphs-3 Workshop and the Decembrettes-6 Conference for their comments, suggestions and feedback on earlier versions of this paper. All errors are mine.

\nocite{hathout2008.textgraphs3}

% Use the \cascadillabibliography command to generate your bibliography. It
% takes the name of your .bib database as its argument. You do not need to
% specify a bibliographystyle; it automatically uses cascadilla.bst, which
% produces `Author (year)' entries.

%\bibliographystyle{cascadilla}
%\bibliography{biblio}

\cascadillabibliography{biblio}

\end{document}